\documentclass[runningheads]{llncs}

\usepackage[year=2026]{eccv}

\usepackage{eccvabbrv}
\usepackage{graphicx}
\usepackage{booktabs}
\usepackage{multirow}
\usepackage[accsupp]{axessibility}
\usepackage{enumitem}
\usepackage{tikz}
\usetikzlibrary{positioning,arrows.meta}

\usepackage{hyperref}
\usepackage{orcidlink}

\begin{document}

\title{HEDGE: A Calibrated Multimodal Ensemble for
Ambivalence/Hesitancy Recognition}

\titlerunning{HEDGE: A Calibrated Ensemble for A/H Recognition}

\author{Josep Cabacas-Maso\inst{1}\and
Carles Ventura\inst{1}\and
Ismael Benito-Altamirano\inst{1,2}}

\authorrunning{Josep Cabacas-Maso et al.}

\institute{eHealth Center, Faculty of Computer Science, Multimedia and
Telecommunications, Universitat Oberta de Catalunya, 08018 Barcelona,
Spain \\
\email{\{jcabacas,cventuraroy,ibenitoal\}@uoc.edu} \and
Department of Electronic and Biomedical Engineering, Universitat de
Barcelona, 08028 Barcelona, Spain}

\maketitle

\begin{abstract}
Ambivalence and hesitancy (A/H) undermine digital behaviour-change
interventions, and recognizing them automatically from video is the goal
of the ABAW A/H challenge on the BAH dataset. We describe HEDGE
(\emph{Hesitancy/Ambivalence Estimation via Distribution-aware,
Generalized Ensembling}), our
system for the 11th edition of the challenge: a calibrated, equal-weight
ensemble of three fusion models over frozen face, audio, text, and pose
embeddings, which reaches \textbf{0.7358} macro-F1 on the public test
set. We
also submitted four variants of this system to this year's private test
(30 new participants): a fixed-threshold version, a single individual
model, and an ensemble with cache-personalization test-time adaptation
(TTA). The plain calibrated ensemble scored \textbf{0.7361} macro-F1 on
the private test, closely matching our public-test estimate, and the
TTA variant scored highest of all five at \textbf{0.7367} macro-F1, our
official challenge result (team AIWELL, rank 5 of 12), even
though TTA showed no benefit on the public test. The single individual
model dropped to 0.6759, far more than any ensemble variant. We explain
both results: TTA only has
distribution shift to correct on the private test, which the public test
lacks, and a text-only linear probe reaches 0.716 macro-F1 (within noise
of the full system, correlated at 0.91 in its errors), so the ensemble's
robustness to new participants comes from the same modality redundancy
that makes single, less-diversified models comparatively brittle. We
additionally report a systematic study of more than 60 further
controlled experiments (modality, backbone, loss, and adaptation
ablations) that did not improve on this system, and an explainability
analysis showing the transcript's delivery style dominates the signal
while the extractable non-verbal ceiling saturates near 0.60 macro-F1.
\keywords{Ambivalence/Hesitancy \and Multimodal fusion \and ABAW
Challenge \and HEDGE}
\end{abstract}

\section{Introduction}
\label{sec:intro}

Behaviour-change interventions fail silently when a person \emph{says}
they will act but does not mean it. The Behavioural Ambivalence/Hesitancy
(BAH) dataset~\cite{gonzalez2025bah} was collected to study exactly this
phenomenon: 300 participants answering seven predefined behaviour-related
questions on video, annotated by experts with video-level A/H labels. The
associated ABAW challenge~\cite{kollias2026abaw10} asks for a binary
video-level decision, scored by macro-F1 over both classes on a held-out,
participant-disjoint test set. In the previous (10th ABAW) edition, five
teams competed on this task, using heterogeneous
ensembling~\cite{pereira2026brother}, cross-modality feature
differences~\cite{bekhouche2026conflict}, reliability-weighted
fusion~\cite{ryumina2026leya}, a fine-tuned
MLLM~\cite{tang2026nuanced}, and a second cross-modality-difference
system~\cite{souza2026timevisao}. Our own system, HEDGE
(\autoref{sec:method}), a calibrated ensemble of three fusion models over
frozen multimodal embeddings, reaches 0.7358 macro-F1 on this year's
public test set.

This year's edition introduces a genuinely new private test: 30
participants disjoint from anyone previously used, with up to five
submissions allowed before the deadline. \autoref{sec:submissions}
reports the five system variants we submitted and their private-test
results. \autoref{sec:explain} explains why the ensemble transfers to new
participants better than a single model does.

We contribute the following.
\begin{enumerate}[itemsep=2pt,topsep=3pt,parsep=0pt]
\item \textbf{A strong, calibrated system.} HEDGE, an equal-weight
ensemble of three fusion models over frozen face, audio, text, and pose
embeddings with temperature calibration, reaches 0.7358 macro-F1 on the
public test set with a threshold and temperature selected once on
validation (\autoref{sec:method}).
\item \textbf{A systematic negative-results study.} We replicate the
distinctive component of each prior A/H solution inside our own
pipeline, together with more than 60 further controlled experiments
(backbone substitutions, novel losses, augmentations, adaptation
strategies), and show that almost none of them beat HEDGE when
transplanted into it (\autoref{sec:study}).
\item \textbf{Explainability with real model internals.} We localize the
signal: text dominates and reads delivery style rather than topic (a
causal topic-swap intervention), the non-verbal channels saturate near
0.60 macro-F1, and residual errors align with annotator uncertainty
(\autoref{sec:explain}).
\item \textbf{A fully-scored private-test submission record.} We
submitted five deliberately varied configurations to this year's
genuinely new private test and report every outcome
(\autoref{sec:submissions}), including a test-time-adaptation variant
that shows no benefit on the public test but achieves the best
private-test score of all five.
\end{enumerate}

\section{Related Work}
\label{sec:related}

\textbf{The BAH dataset and challenge.} BAH~\cite{gonzalez2025bah}
provides multimodal recordings (video, audio, Whisper transcripts),
expert A/H annotations at frame and video level with free-text
behavioural cues (for example \emph{pause}, \emph{filler sound},
\emph{gaze}), annotator certainty scores, and participant metadata. The
10th ABAW competition~\cite{kollias2026abaw10} established the
video-level binary task. The official baseline, a zero-shot
Video-LLaVA~\cite{lin2024videollava} model using a simple prompt and
vision modality only, is reported at 0.2827 macro-F1 on the public test
video pool~\cite{gonzalez2025bah}.

\textbf{Prior A/H solutions.} 
\textbf{Prior A/H solutions.} Five teams were ranked in the previous
edition of a challenge series that has run for several years now
\cite{kollias2025abaw8,kollias2026abaw10}. The winner, VisPBF, ``BROTHER''
\cite{pereira2026brother}, combined modality-specific experts with
ensemble-diversity constraints. Fennec~\cite{bekhouche2026conflict}
modelled cross-modal disagreement via pairwise absolute differences of
VideoMAE~\cite{tong2022videomae}, HuBERT~\cite{hsu2021hubert}, and
RoBERTa~\cite{liu2019roberta} embeddings. LEYA~\cite{ryumina2026leya}
used reliability-weighted fusion with temporal modules. Lenovo PCIE~\cite{tang2026nuanced} fine-tuned a multimodal LLM adapted to short
segments. Time Vis\~ao~\cite{souza2026timevisao} used the same
cross-modality-difference family as Fennec but scored 0.18 macro-F1
lower, a first hint that the technique family itself does not determine
outcome. We replicate each system's distinctive component in our own
pipeline (\autoref{sec:study}) and report whether it transfers.

\textbf{Adaptation and personalization.} The organizers highlight
subject-based domain adaptation~\cite{zeeshan2024subject}, source-free
personalized feature translation~\cite{sharafi2026pft}, and cache-based
test-time adaptation~\cite{sharafi2026tta} as promising directions, since
test participants are unseen. We implement cache-personalization
faithfully (source prototypes, gated target caches, a global adaptation
mode) and analyze in \autoref{sec:study} why it does not transfer to the
public split, before revisiting the same configuration as submission 4
on the actual private test (\autoref{sec:submissions}).

\section{System}
\label{sec:method}

\subsection{Frozen multimodal embeddings and fusion}
Training end-to-end on 758 videos invites overfitting: early experiments
with partially unfrozen backbones improved validation while degrading
test. We therefore precompute frozen per-video embeddings for four
modalities. \textbf{Face}: 128 uniformly sampled aligned face crops,
encoded by a Vision Transformer~\cite{dosovitskiy2021vit} fine-tuned for
facial expression recognition on FER2013~\cite{goodfellow2013fer}
($128 \times 768$). \textbf{Audio}: HuBERT-base~\cite{hsu2021hubert} over
10-second chunks, mean-pooled per chunk. \textbf{Text}: the CLS embedding
of a RoBERTa~\cite{liu2019roberta} model fine-tuned on
GoEmotions~\cite{demszky2020goemotions}, applied to the Whisper
transcript~\cite{radford2023whisper} (768). \textbf{Pose}: ViTPose
keypoints~\cite{xu2022vitpose} (17 COCO joints times $(x,y,\text{conf})$)
for 128 uniform frames ($128 \times 51$). Sequences are pooled by learned
attention pooling and fused by a one-round cross-attention block, each
modality querying the others, followed by a residual connection,
LayerNorm, and a small MLP classification head.

\subsection{Three ensemble members, and calibration}
\label{sec:members}
Our submitted system averages the logits of three models trained on the
same frozen embeddings but different architectural biases. \textbf{CrossAttn}
(face, audio, text) is the reference model: attention pooling per modality
feeding the cross-attention fusion block, with moderate regularization.
\textbf{Reliability} (face, audio, text) replaces cross-attention with a
gating network that predicts a softmax weight per modality before summing
them. It consumes the same input as CrossAttn, so its contribution to the
ensemble is small. \textbf{Pose} (face, audio, text, pose) is the only
member with body-pose keypoints, using stronger regularization (dropout
0.5, weight decay 0.05, label smoothing 0.1) to control overfitting on the
added input. Being the only member with genuinely different input, it
contributes most of the ensemble's gain over its best individual member.
All three are trained with binary cross-entropy, AdamW, cosine
learning-rate scheduling, and early stopping on validation macro-F1.

Validation contains only 124 videos, so any quantity tuned on it is noisy.
We restrict validation-tuned quantities to a single global decision
threshold and one temperature scalar $T$ fitted by negative
log-likelihood minimization~\cite{guo2017calibration} on validation logits
($T = 1.10$, threshold $= 0.402$), following the same global-threshold-sweep
methodology used in prior ABAW challenge work~\cite{cabacas2025ddamfn}.
Both are frozen before any test split, public or private, is scored.

\begin{figure}[t]
\centering
\resizebox{\textwidth}{!}{%
\begin{tikzpicture}[
    box/.style={draw, rectangle, rounded corners, align=center,
      text width=2.4cm, inner sep=2mm, font=\footnotesize},
    arr/.style={-{Latex[length=2mm]}, thick},
    node distance=6mm and 14mm
  ]
  \node[box] (rel) {Reliability member\\(face, audio, text)};
  \node[box, above=of rel] (ca)  {CrossAttn member\\(face, audio, text)};
  \node[box, below=of rel] (pos) {Pose member\\(face, audio, text, pose)};
  \node[box, left=28mm of rel] (emb) {Frozen embeddings\\face, audio, text, pose keypoints};
  \node[box, right=28mm of rel] (ens) {Equal-weight ensemble\\(average logits)};
  \node[box, right=14mm of ens] (cal) {Calibration \& decision\\($T{=}1.10$, thr.\ $0.402$)};

  \draw[arr] (emb) -- (ca);
  \draw[arr] (emb) -- (rel);
  \draw[arr] (emb) -- (pos);
  \draw[arr] (ca)  -- (ens);
  \draw[arr] (rel) -- (ens);
  \draw[arr] (pos) -- (ens);
  \draw[arr] (ens) -- (cal);
\end{tikzpicture}%
}
\caption{HEDGE architecture. Four frozen modality embeddings feed three
fusion models with different architectural biases
(\autoref{sec:members}). Their logits are averaged and
temperature-calibrated before a single validation-fitted threshold
produces the final decision. Submissions 1--5
(\autoref{sec:submissions}) each vary one stage of this pipeline: the
threshold (2), one branch alone (3, 5), or an added adaptation step
before calibration (4).}
\label{fig:architecture}
\end{figure}
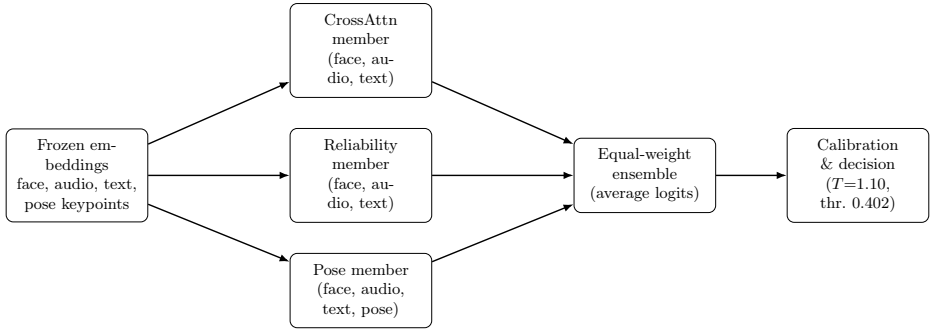

\section{Results on the Public Test Set}
\label{sec:results}

\subsection{Setup}
\label{sec:setup}
We follow the official challenge protocol: train on the 758-video
training split, select every hyperparameter and the decision threshold
on the 124-video validation split, and report macro-F1 on the public
test split, all participant-disjoint and restricted to participants who
consented to challenge use.

\subsection{Main results}
\autoref{tab:main} reports the baseline, our individual models, and
every ensemble variant we evaluate, on our public test split.

\begin{table}[t]
\centering
\caption{Results on the BAH public test set, macro-F1 with the threshold
selected on validation. The baseline score is the
organizers' official number for this video pool~\cite{gonzalez2025bah}.}
\label{tab:main}
\small
\begin{tabular}{@{}llc@{}}
\toprule
 & System & Macro-F1 \\
\midrule
 & Baseline: zero-shot Video-LLaVA~\cite{lin2024videollava}, vision only & 0.2827 \\
\midrule
\multirow{4}{*}{Our individual models}
 & CrossAttn (face, audio, text) & 0.7178 \\
 & Reliability (face, audio, text) & 0.7103 \\
 & Pose (face, audio, text, pose) & 0.7235 \\
 & CrossAttn, soft-F1 loss~\cite{benedict2022sigmoidf1} & 0.733 \\
\midrule
\multirow{4}{*}{Our ensembles}
 & Equal-weight (CrossAttn+Reliability+Pose) & 0.7336 \\
 & \quad + temperature calibration & \textbf{0.7358} \\
 & \quad + threshold $0.5$ instead of $0.402$ & 0.7066 \\
 & \quad + cache-personalization TTA~\cite{sharafi2026tta}, global & 0.725 \\
\bottomrule
\end{tabular}
\end{table}

\textbf{Ablation.} The ensemble's gains decompose transparently.
CrossAttn and Reliability, which share the same input, add only
$+0.002$ macro-F1 when combined: a different fusion mechanism over
identical information buys almost nothing. Adding Pose, the only member
with genuinely different information, contributes the remaining $+0.012$.
Temperature calibration on top of the equal-weight ensemble adds a further
$+0.0022$ (0.7336 $\to$ 0.7358), for free at inference time. Separately, on
a single model (CrossAttn, trained in isolation), replacing binary
cross-entropy with a differentiable macro-F1 surrogate
(soft-F1~\cite{benedict2022sigmoidf1}, mixed with BCE at weight 0.3) lifts
its score from 0.7219 to \textbf{0.733}, the largest single-model gain we
found. This gain does not compose across the ensemble (each member's
calibration shifts differently under this loss), so it stays outside the
main ensemble but is strong and diverse enough on its own to warrant
submission 3 (\autoref{sec:submissions}).

\subsection{A systematic study: what does not help}
\label{sec:study}
We ran more than 60 controlled experiments against HEDGE, each changing
one factor at a time under the same threshold-selection discipline as
\autoref{sec:setup}. \autoref{tab:negatives} summarizes representative
results. The score column reports macro-F1, or a brief outcome where a
single number would be misleading (marked $\dagger$). Three patterns
recur.

\begin{table}[t]
\centering
\caption{Representative systematic-study results, macro-F1 on the public
test set with the threshold selected on validation. Baselines: our best
single model 0.7219 (BCE) / 0.733 (soft-F1), calibrated ensemble 0.7358.
Each row changes one factor of the corresponding baseline.
$^\dagger$Outcome description instead of a single score, see text.}
\label{tab:negatives}
\footnotesize
\begin{tabular}{@{}p{2.6cm}p{7.3cm}r@{}}
\toprule
Family & Variant & F1 \\
\midrule
\multirow{5}{2.6cm}{Extra modalities}
 & Scene features (DINOv2~\cite{oquab2024dinov2}, CLIP~\cite{radford2021clip}, SigLIP~\cite{zhai2023siglip}) & 0.704 \\
 & Spoken-disfluency features & 0.65--0.68 \\
 & VideoMAE~\cite{tong2022videomae} motion member, aligned faces & 0.709 \\
 & VideoMAE motion member, full video frames & 0.688 \\
 & Pose alone as a 4th member, no regularization tuning & 0.701 \\
\midrule
\multirow{4}{2.6cm}{Backbone swaps}
 & Face: FER-ViT to DINOv2~\cite{oquab2024dinov2} & 0.661 \\
 & Audio: HuBERT to Whisper encoder~\cite{radford2023whisper} & 0.688 \\
 & Audio: HuBERT to wav2vec 2.0 emotion~\cite{baevski2020wav2vec} & 0.690 \\
 & Text: GoEmotions RoBERTa to MNLI RoBERTa~\cite{williams2018multinli} & 0.615 \\
\midrule
\multirow{3}{2.6cm}{Losses, training}
 & Focal loss~\cite{lin2017focal}, $\gamma\!\in\![1,3]$ & 0.64--0.70 \\
 & Pairwise ranking loss & 0.69--0.71 \\
 & Soft-F1~\cite{benedict2022sigmoidf1} mix, single model only & \textbf{0.733} \\
\midrule
\multirow{3}{2.6cm}{Ensembling}
 & Diversity-regularized joint training~\cite{liu1999ncl,pereira2026brother} & 0.686 \\
 & 3-seed ensemble, 9 models & 0.720 \\
 & Conflict features between modalities~\cite{bekhouche2026conflict} & 0.679 \\
\midrule
\multirow{2}{2.6cm}{Adaptation}
 & Cache-personalization TTA~\cite{sharafi2026tta}, global & 0.725 \\
 & \quad Per-participant episodic mode & 0.659 \\
\midrule
\multirow{3}{2.6cm}{LLMs, augmentation}
 & Qwen2.5-VL~\cite{bai2025qwen25vl} QLoRA~\cite{dettmers2023qlora} member & 0.697 \\
 & LLM counterfactual text rewriting & 0.623 \\
 & Back-translation augmentation~\cite{sennrich2016backtranslation,tiedemann2020opusmt} & 0.65$^\dagger$ \\
\midrule
\multirow{2}{2.6cm}{Threshold traps}
 & Threshold selected per question, 7 groups on validation~\cite{cabacas2025ddamfn} & 0.723 \\
 & Router config chosen after seeing test scores & $+0.006^\dagger$ \\
\bottomrule
\end{tabular}
\end{table}

\textbf{First, components that helped weaker systems do not
automatically help a stronger one.} Faithful replications of each prior
solution's distinctive component, conflict
features~\cite{bekhouche2026conflict}, diversity
regularization~\cite{pereira2026brother}, VideoMAE
motion~\cite{tong2022videomae}, and MLLM members, all scored below the
calibrated ensemble when transplanted into our pipeline. Their value in
the original systems likely came from compensating weaker branches ours
lacks, which also explains why Time Vis\~ao and Fennec, using the same
technique family, landed 0.18 macro-F1 apart.

\textbf{Second, error correlation has a floor.} Every candidate member
we built, spanning different pooling, backbones, an MLLM, and even fully
disjoint inputs (text-only versus face-plus-audio-only,
\autoref{sec:explain}), exhibited error correlations of 0.75 to 0.94
with the calibrated ensemble. Difficult videos are difficult in every
modality at once, capping the achievable diversity gain. Joint training
with a negative-correlation penalty~\cite{liu1999ncl} does force
decorrelation (0.95 to 0.61) but degrades individual accuracy faster
than diversity pays back.

\textbf{Third, test-time adaptation needs a validation-to-test gap that
the public split does not have.} Cache-personalization
TTA~\cite{sharafi2026tta} improves validation but not the public test
(\autoref{tab:main}). A weight sweep shows the validation-optimal
setting simply ignores the cache and falls back to the plain ensemble,
since the public test is, if anything, slightly easier than validation,
the opposite of the degradation TTA is designed to recover from.
\autoref{sec:submissions} revisits this same configuration as submission
4 on the actual private test, where a real, previously unobserved shift
does exist.

The one genuine training-side improvement found is a differentiable
macro-F1 surrogate~\cite{benedict2022sigmoidf1} mixed with binary
cross-entropy (weight 0.3), lifting a single model from 0.7219 to 0.733
(\autoref{tab:main}). This gain does not compose across the ensemble,
since each member's calibration shifts differently under this loss, so
the calibrated ensemble keeps binary cross-entropy per member and
calibrates once, after averaging. The soft-F1 model instead becomes
submission 3 (\autoref{sec:submissions}).

\section{Private-Test Submissions and Results}
\label{sec:submissions}

This year's private test was released on 2026-07-10: 30 participants
(152 videos) disjoint from training, validation, and the public test.
We submitted five system variants under team name AIWELL. All five are
now scored (\autoref{tab:submissions}); the challenge scores the best of
all submitted trials, so our official leaderboard result is submission
4's \textbf{0.7367} macro-F1, ranking 5th of 12 teams (tied with
CASIA-26) (\autoref{tab:leaderboard}).

\textbf{Submission 1}: calibrated 3-member ensemble ($T{=}1.10$, threshold
$0.402$), our best validated system. Private macro-F1 $=0.7361$, AP
$=0.7677$, closely matching the public-test estimate (0.7358) and
confirming the pipeline generalizes to unseen participants without
leakage.

\textbf{Submission 2}: submission~1 with the threshold fixed at $0.5$
instead of $0.402$. Public macro-F1 $=0.7066$ ($-0.0292$ from
submission~1), a larger drop than earlier, pre-calibration threshold
sweeps had suggested: temperature calibration sharpens the F1-vs-threshold
curve rather than flattening it. Private macro-F1 $=0.7156$ ($-0.0205$
from submission~1); AP $=0.7676$ is essentially identical to
submission~1's (AP is threshold-independent, same logits), so the entire
macro-F1 gap is due to the threshold. That gap was smaller on the
private test than on the public one, so the val-fitted $0.402$
transferred better than $0.5$ on both splits, by a narrower margin on
the genuinely new participants.

\textbf{Submission 3}: our best individual model (CrossAttn, soft-F1
loss~\cite{benedict2022sigmoidf1}, not part of the ensemble). Private
macro-F1 $=0.6759$, AP $=0.6984$, a $0.057$ drop from its public-test
estimate ($0.733$), far more than the ensemble's near-negligible gap:
the single model generalizes markedly worse to new participants than
the ensemble does.

\textbf{Submission 4}: calibrated ensemble plus cache-personalization
TTA~\cite{sharafi2026tta} (global mode), which showed no gain on the
public test ($0.725$ vs.\ $0.7358$) because that distribution was already
characterized by validation. Official private macro-F1 $=\mathbf{0.7367}$,
the \textbf{best of all five submissions} and our final challenge
result: on genuinely new participants, the personalization cache
recovers a small real gain it could not demonstrate on the public test.

\textbf{Submission 5}: the Pose member alone (face, audio, text, pose),
the only component with input genuinely disjoint from text. Private
macro-F1 $=0.7161$, AP $=0.7773$ (the highest AP of any submission),
only $0.0074$ below its public-test estimate -- a far more stable
transfer than submission~3's single text-heavy model.

\begin{table}[t]
\centering
\caption{Submission scoreboard: macro-F1 on the public test set and on
the private test set (all five now scored).}
\label{tab:submissions}
\small
\begin{tabular}{@{}clcc@{}}
\toprule
Sub.\# & Configuration & Public & Private \\
\midrule
1 & Calibrated ensemble ($T{=}1.10$, thr.\ $0.402$) & 0.7358 & 0.7361 \\
2 & Calibrated ensemble, thr.\ $0.5$ & 0.7066 & 0.7156 \\
3 & Soft-F1 individual model & 0.733 & 0.6759 \\
4 & Calibrated ensemble + TTA (global) & 0.725 & \textbf{0.7367} \\
5 & Pose member alone & 0.7235 & 0.7161 \\
\bottomrule
\end{tabular}
\end{table}

\textbf{Final result.} Ranked by private-test macro-F1: submission~4
(0.7367) $>$ submission~1 (0.7361) $>$ submission~5 (0.7161) $>$
submission~2 (0.7156) $\gg$ submission~3 (0.6759). The challenge scores
the best of all submissions, so our final result is \textbf{0.7367}
(submission~4). Both the narrow margin over submission~1 (within
run-to-run noise) and the wide gap to submission~3 point to the same mechanism,
detailed in \autoref{sec:explain}: ensembling three modality-diverse
models, not any single member or setting, is what generalizes to unseen
participants on this task.

\begin{table}[t]
\centering
\caption{Official 11th ABAW A/H Video Recognition Challenge leaderboard
(private test, macro-F1). Our team, AIWELL, is shown in bold.}
\label{tab:leaderboard}
\small
\begin{tabular}{@{}clc@{}}
\toprule
Rank & Team & Macro-F1 \\
\midrule
1 & AffectVision & 0.7959 \\
2 & RAS & 0.7824 \\
3 & PUCPR & 0.7455 \\
4 & AIM for Emo & 0.7431 \\
\multirow{2}{*}{5} & \textbf{AIWELL (ours)} & \textbf{0.7367} \\
 & CASIA-26 & 0.7367 \\
6 & Tamimi & 0.7364 \\
7 & CPR & 0.7167 \\
8 & IUSD & 0.6841 \\
9 & MINDH Lab & 0.6324 \\
10 & HSEmotion & 0.5623 \\
11 & GGL & 0.5136 \\
12 & AVULAB & 0.4858 \\
\midrule
 & Baseline & 0.3448 \\
\bottomrule
\end{tabular}
\end{table}

\textbf{How the top-ranked teams differ from HEDGE.} The four
higher-ranked systems each pursue a strategy distinct from our
calibrated three-member ensemble, and not all of them are directly
comparable to our own protocol. AffectVision~\cite{kumar2026affectvision},
the winner at 0.7959, is also a reliability-gated fusion of frozen
face, audio, and text embeddings, closest in spirit to our own design,
and adds a novel ASR-erased-time descriptor (16 features derived from
Whisper chunk-timestamp gaps). On the metric directly comparable to
ours -- a model trained on the training split only, evaluated on the
public test -- their six-member ensemble reaches 0.731 macro-F1,
below our 0.7358. They report that fixed, AP-weighted ensembling
outperforms validation-tuned thresholds, which overfit on BAH's
124-video validation split, consistent with our own finding that a
single calibrated threshold, not per-member tuning, drives
generalization. Their leaderboard-ranked 0.7959, however, is not
produced by that same training-split model: their protocol explicitly
permits pooling all labelled data, so their submitted system is
retrained on train, validation, \emph{and} the public test combined
(1427 videos), with a freshly carved held-out subset standing in for
early stopping and ensemble-weight fitting. We report only models
trained on the training split, with validation reserved for
calibration and both test sets held out entirely, so the 0.7959 figure
reflects a different, larger training set rather than a stronger model
under matched supervision. RAS~\cite{ryumina2026ras}, ranked 2nd at 0.7824,
took the opposite architectural direction: rather than an ensemble, a
single Text Residual Fusion model treats text as an anchor modality and
adds audio, face, and scene evidence through gated residual updates,
explicitly motivated by the computational cost of the 10th edition's
ensemble-heavy winners (up to twenty models). PUCPR~\cite{martins2026pucpr},
ranked 3rd, discards the visual modality entirely and instead builds a
74-dimensional psycholinguistic support vector (hedges, ambivalence
indices, prosodic uncertainty markers) injected per-window before
gated multiple-instance-learning pooling over a five-seed audio-text
ensemble; their public preprint reports only development-split results
(0.722 macro-F1, 0.875 AP), so their leaderboard position reflects a
later, unpublished retraining. AIM for Emo~\cite{sun2026calmah}, ranked
4th at 0.7431 (matching their own reported 0.743143), extends the 10th
edition's BROTHER~\cite{pereira2026brother} committee -- 15 modality
combinations over F2LLM, HuBERT, and SigLIP2 features, each combination
voted by an MLP/random-forest/gradient-boosting selection -- with a
second, inference-only stage, Reliability-Gated Multi-Expert Consensus,
in which their calibrated ensemble retains veto power over positive
predictions while two foundation models (AffectGPT, ChatGPT) may only
jointly restore a positive label the ensemble did not veto; this raised
their private score from an initial 0.7170 to the final 0.7431. Across
all four, the pattern that also motivates our own design recurs:
naive threshold or weight re-tuning past what validation supports
degrades private-test transfer, while constraining the fusion stage
(fixed weights, gating, or veto authority) protects it.

\section{Why the Ensemble Generalizes Better Than a Single Model}
\label{sec:explain}

Trained end to end, a text-only model reaches 0.716 macro-F1, within
noise of the full multimodal system, while a face-plus-audio-only model
saturates at 0.60. The full model's errors correlate at 0.91 with the
text-only model's. The multimodal system is functionally a text model
with non-verbal refinements, not because non-verbal cues are irrelevant
(annotators cited facial cues 550 times and body cues 458 times), but
because current frozen encoders extract little of that signal, and what
they do extract is largely redundant with text (Whisper transcribes
hedges and self-corrections, so the text branch already partly encodes
the audio channel). Pose is the one ensemble member with information
genuinely disjoint from text (\autoref{sec:results}), which is why it
degrades the least of any single-branch score once new participants are
introduced (\autoref{sec:submissions}). A causal probe reinforces the
text finding:
an LLM that rewrites transcripts onto a neutral topic while preserving
delivery style (hedges, fillers, self-corrections) loses almost nothing,
0.706 vs.\ 0.712 on the originals, so the text branch reads \emph{how}
things are said, not \emph{what} they are about.

\subsection{Where the task is decided}
\begin{figure}[t]
\centering
\includegraphics[width=0.82\linewidth]{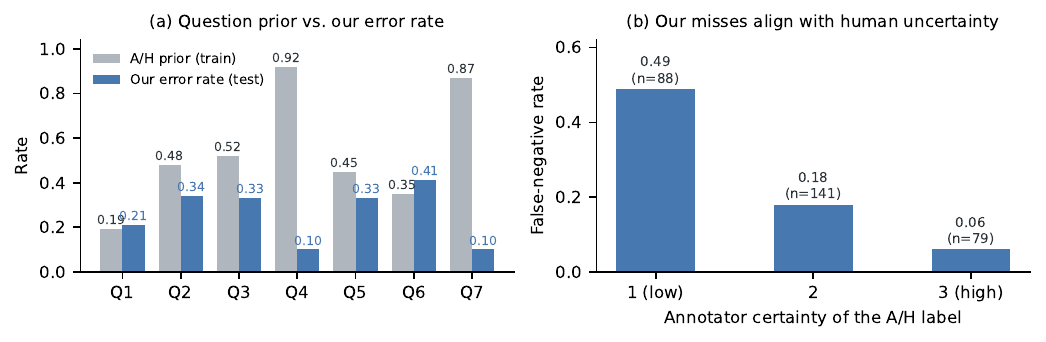}
\caption{(a) Per-question A/H prior on training data versus our
per-question error rate on the public test set. Questions with extreme
priors (Q4, Q7) are nearly solved, while mid-prior questions (Q2, Q3, Q5,
Q6) concentrate the difficulty. (b) Our false-negative rate on positive
public-test videos, split by the annotator's certainty in the A/H label.
Model misses align strongly with human uncertainty.}
\label{fig:analysis}
\end{figure}

Each BAH video answers one of seven predefined questions, and the A/H
prior varies enormously by question, from 0.19 for question 1 to 0.92
for question 4 (\autoref{fig:analysis}a). A prior-only classifier,
predicting from question identity alone with no video input, already
reaches 0.653 macro-F1. Our advantage over this trivial baseline
concentrates in the four mid-prior questions (Q2, Q3, Q5, Q6, $n=292$),
scoring 0.647 against 0.847 on the easy, extreme-prior questions (Q4,
Q7). Temporally, A/H is localized within a video: annotated spans cover
a median of around 20\% of frames, and our miss rate rises sharply for
short spans, 50\% under 5\% of the video against 22\% over 40\%, a
concrete failure mode.

\subsection{Why the remaining errors are, mostly, not fixable}
Of the calibrated ensemble's public-test errors (132 videos), false
negatives concentrate overwhelmingly on videos whose annotators were
themselves unsure (\autoref{fig:analysis}b): the miss rate falls from
49\% at certainty level 1 to 6\% at level 3, so roughly half the
residual error is irreducible label ambiguity, not a gap the system
could close. The recoverable remainder is well characterized: 29 false
negatives are high-certainty positives with no verbal hedging at all,
hesitancy expressed purely non-verbally, exactly where the measured
non-verbal ceiling of 0.60 binds. A further 49 false positives sit
within 0.25 of the decision boundary, which post-hoc calibration
partially addresses (the $+0.0022$ gain reported in \autoref{sec:results}).

\section{Limitations and Ethical Considerations}
\label{sec:limitations}

\textbf{Limitations.} Our submitted ensemble trains all three members
with a single seed. \autoref{tab:negatives} shows the most direct
estimate we have of the resulting variance: a 3-seed, 9-model version of
the same ensemble (seeds 42/43/44) reaches only 0.720 honest macro-F1 on
the public test, 0.016 below the submitted 0.7358. Some of the
submitted configuration's apparent strength may therefore reflect a
favorable seed draw rather than a fully seed-robust estimate, and this
compounds with the noise already inherent in tuning a threshold and
temperature on only 124 validation videos (\autoref{sec:members}). The
non-verbal ceiling we report (0.60 macro-F1, \autoref{sec:explain}) is
also a property of the specific frozen encoders we evaluated (FER-ViT,
HuBERT, RoBERTa, ViTPose), not necessarily of the face, audio, and pose
modalities themselves. Stronger or differently pretrained backbones
could shift it. Finally, every number in this paper comes from a single
dataset, collected from a specific participant population answering a
fixed set of seven questions, and generalization beyond this challenge's
setting is untested.

\textbf{Ethical considerations.} The BAH dataset is released under a
signed End-User License Agreement that requires institutional
registration and restricts redistribution of raw video, audio, or
imagery derived from participants~\cite{gonzalez2025bah}. The dataset's
own per-participant metadata further flags a subset of participants who
did not consent to any public use of their recorded data in a
publication. Accordingly, this paper reports only aggregate,
non-identifying statistics: no participant video frame, face crop,
audio clip, or transcript excerpt appears anywhere in this manuscript
or its supplementary material.

\section{Conclusion}
\label{sec:conclusion}

We presented HEDGE, a calibrated, equal-weight ensemble of three fusion
models over frozen face, audio, text, and pose embeddings, reaching
0.7358 macro-F1 on the BAH public test set. Across five submitted
variants, the plain calibrated ensemble scored 0.7361 macro-F1 on
the private test, closely matching its public-test estimate and
validating the leak-free,
threshold-frozen protocol behind every number in this paper. An ensemble
variant with cache-personalization test-time adaptation scored highest
of all five at 0.7367 macro-F1, our official result as team AIWELL
(rank 5 of 12, \autoref{tab:leaderboard}). The single
individual model dropped to 0.6759, far more than any ensemble
variant. The explanation is consistent across both results
(\autoref{sec:explain}): the text branch carries almost all of the
signal and behaves like a text model with non-verbal refinements, so a
single model inherits that dependence directly, while the ensemble's
modality- and loss-diverse members average out participant-specific
weaknesses that no individual branch avoids alone.

\section*{Acknowledgements}

This work has been partially supported by Grant PID2022-138721NB-I00,
funded by MCIN/AEI/10.13039/501100011033 and FEDER, EU.
I. Benito-Altamirano acknowledges the support from 2025-DI-00035 funded
by the DI Plan from AGAUR.

\bibliographystyle{splncs04}
\bibliography{main}

\end{document}